\RequirePackage{fix-cm}
\documentclass[twocolumn]{svjour3}          
\smartqed  

\usepackage{graphicx}
\usepackage{amsmath}
\usepackage{amssymb}

 \usepackage{cite}

\graphicspath{{Images/}}
\DeclareGraphicsExtensions{.pdf,.jpeg,.png,.jpg,.svg}
\usepackage{multicol}
\usepackage{textcomp}
\usepackage{makecell}
\usepackage{multirow}
\usepackage{multicol}
\usepackage{url}
\usepackage{xcolor}
\usepackage{subfig}

\definecolor{light-gray}{gray}{0.7}
\usepackage[switch]{lineno}

\journalname{Preprint accepted in Springer Knowledge and Information Systems (KAIS)}
\begin{document}

\title{Semantic Driven Multi-Camera Pedestrian Detection}


\author{Alejandro L\'{o}pez-Cifuentes, Marcos Escudero-Vi\~nolo, Jes\'{u}s~Besc\'{o}s, and Pablo Carballeira}

\institute{Alejandro L\'{o}pez-Cifuentes (Corresponding Author) \at Universidad Aut\'{o}noma de Madrid. Video Processing and Understanding Lab (VPULab).    \email{alejandro.lopezc@uam.es}    \and Marcos Escudero-Vi\~nolo \at Universidad Aut\'{o}noma de Madrid. Video Processing and Understanding Lab (VPULab). Email: marcos.escudero@uam.es \and Jes\'{u}s~Besc\'{o}s \at Universidad Aut\'{o}noma de Madrid.  Video Processing and Understanding Lab (VPULab). E-mail: j.bescos@uam.es \and Pablo Carballeira \at Universidad Aut\'{o}noma de Madrid. Video Processing and Understanding Lab (VPULab). E-mail: pablo.carballeira@uam.es}


\authorrunning{L\'{o}pez-Cifuentes et. al.}
\maketitle

\begin{abstract}
In the current worldwide situation, pedestrian detection has reemerged as a pivotal tool for intelligent video-based systems aiming to solve tasks such as pedestrian tracking, social distancing monitoring or pedestrian mass counting. Pedestrian detection methods, even the top performing ones, are highly sensitive to occlusions among pedestrians, which dramatically degrades their performance in crowded scenarios. The generalization of multi-camera set-ups permits to better confront occlusions by combining information from different viewpoints. In this paper, we present a multi-camera approach to globally combine pedestrian detections leveraging automatically extracted scene context. Contrarily to the majority of the methods of the state-of-the-art, the proposed approach is scene-agnostic, not requiring a tailored adaptation to the target scenario\textemdash e.g., via fine-tunning. This noteworthy attribute does not require \textit{ad hoc} training with labelled data, expediting the deployment of the proposed method in real-world situations. Context information, obtained via semantic segmentation, is used 1) to automatically generate a common Area of Interest for the scene and all the cameras, avoiding the usual need of manually defining it; and 2) to obtain detections for each camera by solving a global optimization problem that maximizes coherence of detections both in each 2D image and in the 3D scene. This process yields tightly-fitted bounding boxes that circumvent occlusions or miss-detections. Experimental results on five publicly available datasets show that the proposed approach outperforms state-of-the-art multi-camera pedestrian detectors, even some specifically trained on the target scenario, signifying the versatility and robustness of the proposed method without requiring ad-hoc annotations nor human-guided configuration.

\keywords{Pedestrian detection \and multi-camera systems \and semantic segmentation \and video surveillance}

\end{abstract}


\section{Introduction} \label{sec:introduction}
In the current worldwide situation, pedestrian detection has reemerged as a pivotal tool for intelligent video-based systems aiming to solve tasks such as pedestrian tracking, social distancing monitoring or pedestrian mass counting. Automatic people detection is generally considered a solid and mature technology able to operate with nearly human accuracy in generic scenarios \cite{garcia2015people,dollar2012pedestrian, priscilla2019pedestrian}. However, the handling of severe-occlusions is still a major challenge \cite{ning2020survey}. Occlusions occur due to the projection of the 3D objects onto a 2D image plane. Although recent deep-learning based methods are able to cope with partial occlusions, the detection process fails when only a small part or no part of the person is visible. To cope with severe-occlusions, a potential solution is the use of additional cameras: if they are adequately positioned, the different points of view might allow for disambiguation.

\begin{figure}[t!]
	\centering
	\includegraphics[width=0.75\linewidth]{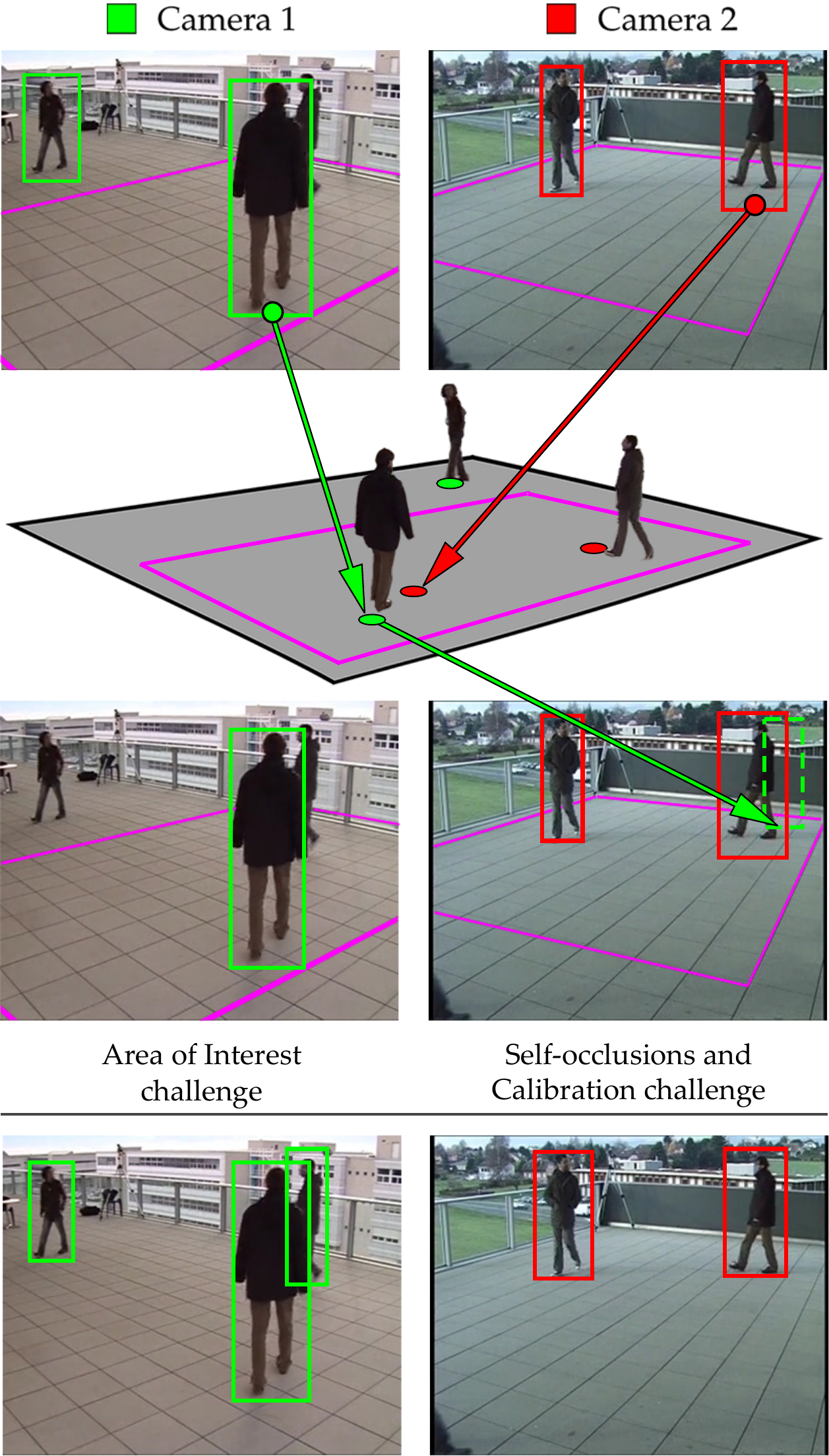}
    \caption{Common challenges of people detection in multi-camera scenarios. First row: Per-camera people detection by Faster-RCNN \cite{renNIPS15fasterrcnn} (solid bounding boxes) with superimposed---in pink, manually annotated Area Of Interest \((\mathcal{AOI})\) from \cite{peng2015robust}. Second row: Reference ground-plane with projected \(\mathcal{AOI}\) and detections (color dots). Area of Interest challenge: one projected detection lays outside of the \(\mathcal{AOI}\) and is filtered-out. Self-occlusions and Calibration challenges: projected detections from different camera views diverge in the common plane due to self-occlusions and calibration errors. The back-projection of a detection from Camera 1 onto Camera 2 creates a miss-aligned bounding box (dotted line). Fourth row: qualitative results of the proposed multi-camera pedestrian detection method using the whole imaged floor as \(\mathcal{AOI}\) and aligned back-projected detections. Better viewed in color.}
	\label{fig:Multi-camera problems}
\end{figure}

Disambiguation is generally achieved by projecting every camera's detections on a common reference plane. The ground plane is usually the preferred option as it constitutes a common reference in which people's height can be disregarded. Per-camera detections can then be combined on the ground plane to refine and complete pedestrian detection. However, there are several challenges to be addressed during this combination or fusion process. Among the striking ones are: the convenience to define common visibility areas where cameras' views overlap, and how to cope with camera calibration errors and persons’ self-occlusions. See Figure \ref{fig:Multi-camera problems} for visual examples of these challenges, which we detail below:

In multi-camera approaches a common strategy is to define an \textbf{operational area} or Area Of Interest \(\mathcal{AOI}\) on the ground plane. This area represents the overlapping field-of-view of all the involved cameras. It can be used to reduce the impact of calibration errors in the process and to generally ease the fusion of per-camera detections. This area is generally manually defined for each scenario, precluding the automation of the process.

\textbf{Scene calibration} is a well-known task \cite{hartley2003multiple} which can be performed either manually or using automatic calibration methods based on image cues. In both cases, small perturbations in the calibration process may cause uncertainty in the fusion of the detections on the ground plane. The impact of calibration errors increases with the distance to the camera: generally, calibration is more accurate for pixels belonging to objects close to the camera.

\textbf{Self-occlusions} are caused by the intrinsic three-dimensional nature of people, resulting in the occlusion of some human parts by some others. If the visible parts are different for different cameras and these are used to project a person location on the ground plane, the cameras' projections will diverge, hindering their fusion.

To cope with these challenges, in this paper we present a multi-camera pedestrian detection method which is driven by semantic information automatically extracted from the 2D images and transferred to the 3D ground plane, and includes the following novel contributions:
\begin{enumerate}
  \item A novel approach to globally combine pedestrian detections in a multi-camera scenario by creating connected components in a graph representation of detections. 
  \item An height-adaptive optimization algorithm which uses semantic cues to globally refine the location and size of people detections by aggregating information from all the cameras.
\end{enumerate}
The proposed method is applied over an operational area in the ground plane, which is automatically defined  by an adaptation of the method described in \cite{lopez2018automatic}.

Experimental results on public datasets (PETS 2009 \cite{web-PETS}, EPFL RLC \cite{baque2017deep,web-RLC}, EPFL Terrace \cite{fleuret2008multicamera,web-Terrace} and EPFL Wildtrack \cite{chavdarova-et-al-2018,web-wildtrackDataset}) prove that the proposed method: (1) outperforms state-of-the-art monocular pedestrian detectors \cite{renNIPS15fasterrcnn,redmon2018yolov3}, (2) outperforms state-of-the art scene-agnostic multi-camera detection approaches; and (3) results in a performance comparable, and even better, to deep-learning multi-camera detection approaches trained and fine-tuned to the target scenario, while not requiring neither a manually annotated operational area nor a specific training on that scenario.

The rest of the paper is organized as follows: Section \ref{Related Work} reviews the State of the Art, Section \ref{Proposed Method} describes the proposed method, Section \ref{Results} presents and discusses experimental results and Section \ref{Conclusions} concludes the paper.


\section{Related Work} \label{Related Work}
Multi-camera people detection faces the combination, fusion and refinement of visual cues from several individual cameras to obtain more people locations.
A common pathway in existing approaches starts by defining an operational area, either manually or, as we propose, based on a semantic segmentation. Then, approaches combining detections using a common reference plane, usually follow a three-stages strategy: (1) extract detections on each camera frame, (2) project detections onto the common plane and (3) combine detections and back-project them to the individual views to obtain per-camera people detections. Finally, obtained detections are sometimes post-processed to further refine their localization.

\subsection{Definition of the operational area} \label{Definition of the operational area}
Some approaches \cite{peng2015robust,utasi2013bayesian} rely on manually annotated operational areas where evaluation is performed. An advantage of these \textit{ad hoc} areas is that the impact of camera calibration errors is limited and controlled. Besides, these areas are defined to maximize the overlapping between the field of view of the involved cameras. However, the manual annotation of these operational areas hinders the generalization of people detection approaches. Our previous work in this domain \cite{lopez2018automatic} resulted in an automatic method for the cooperative extraction of operational areas in scenarios recorded with multiple moving cameras: semantic evidences from different junctures, cameras, and points-of-view are spatio-temporally aligned on a common ground plane and are used to automatically define an operational area or \textit{Area of Interest} (\(\mathcal{AOI}\)).

\subsection{Semantic Segmentation} \label{Semantic Segmentation}
Semantic segmentation is the task of assigning a unique object label to every pixel of an image. During the last years, top performing strategies evolved from the seminal Fully Convolutional Network scheme \cite{long2015fully} and the use of dilated convolutions \cite{YuKoltun2016}. For instance, Zhao et al. \cite{zhao2016pyramid} proposed to implicitly use contextual information by including relationships between different labels---e.g. an airplane is likely to be on a runway or flying in the sky but not on the water. These relationships reduce the inner complexity of datasets with large sets of labels, generally improving performance. Lately, the development and use of new backbones for feature extraction has benefited the task. Zhang et al. \cite{zhang2020resnest} proposed a new ResNet modification called ResNeSt that uses channel-wise attention to capture cross-feature interactions and learn diverse object representations. Similarly, Tao et al. \cite{tao2020hierarchical} proposed the use of a hierarchical attention to combine multi-scale predictions, increasing the performance on the small object instances, as those in PASCAL VOC dataset \cite{everingham2010pascal}.

\subsection{Monocular people detection} \label{Mono-camera people detection}
As stated in Section \ref{sec:introduction}, automatic monocular pedestrian detection is considered a mature technology able to obtain accurate results in a broad range of scenarios. Well established object detectors based on CNNs as Faster-RCNN \cite{renNIPS15fasterrcnn} and YOLOv3 \cite{redmon2018yolov3}, have demonstrated their reliability during the last years. Adapting their core schemes, recent approaches have further increased their performance. Specifically, YOLOv3 has been improved by both decreasing the complexity of the model through new architecture designs \cite{long2020pp} and by efficient model scaling \cite{wang2020scaled}.

Alternatively, novel detectors\textemdash also based on CNNs, have been proposed. Tan et al. \cite{tan2020efficientdet} proposed a weighted bi-directional feature pyramid network allowing easy and fast multi-scale feature fusion, and obtaining a new family of detectors called EfficientDet, that achieved a new state-of-the-art performance in the COCO dataset \cite{lin2014microsoft}. Alternatively, Zhu et al. \cite{zhu2020deformable} proposed to use attention in the form of deformable transformers to also obtain state-of-the-art results.

Nevertheless, even though the most recent works have demonstrated really high performances, in scenarios with severe-occlusions the performance of these algorithms decreases.

\subsection{Projection of per-camera detections} \label{Projection of per-camera detections}
Multi-camera pedestrian detection is fundamentally based on the projection of monocular detections onto a common reference plane. Projection is typically achieved either by using calibrated camera models that relate any 2D image point with a corresponding referenced 3D world direction \cite{utasi2013bayesian}, or by relying on homographic transformations that project image pixels to a specific 3D plane \cite{peng2015robust}. In both cases, the ground plane, where people is usually standing on, is chosen as reference for simplicity reasons.

\subsection{Fusion and refinement of per-camera detections} \label{Fusion}
Fusion and refinement approaches can be mainly divided into three different groups depending on how global detections are obtained. The first group encompasses \textbf{geometrical} methods, which combine detections based on the geometrical intersections between image cues. The second group embraces \textbf{probabilistic} methods, that combine detections via optimization frameworks and statistical modeling of the image cues. The third group is composed of solutions based on the ability of \textbf{deep learning} architectures to model occlusions and achieve accurate pedestrian detection at scene level. 

Regarding \textbf{geometrical} methods, detections are combined by projecting foreground masks to the ground plane in a multi-view scenario: the intersection of foreground regions leads to pedestrian detection \cite{alahi2011sparsity}. Accuracy can be increased by projecting the middle vertical axis of pedestrians, leading to a more accurate intersection on the ground plane and, therefore, to a better estimation of the pedestrian's position \cite{kim2006multi}. Following the same hypothesis,  the use of a space occupancy grid to combine silhouette cues has been proposed: each ground pixel is considered as an occupancy sensor and observations are then used to infer pedestrian detection \cite{franco2005fusion}. All of these approaches outperform single-camera pedestrian detection algorithms by the use of ground-plane homography projections. Nevertheless, the evaluation of foreground intersections in crowded spaces may lead to the appearance of \textit{phantoms} or false detections. To handle this problem, the general multi-camera homography framework has been extended by using additional parallel planes to the ground plane \cite{delannay2009detection,khan2009tracking}. The intersection of the image cues with these parallel planes is expected to suppress these \textit{phantoms}. Similarly, parallel planes can be also used to create a full 3D reconstruction of pedestrians, that can then be back-projected to each of the camera views, improving monocular pedestrian detection \cite{Aliakbarpour2016}. Finally, Lima et al.\cite{lima2021generalizable} replicates a preliminar version of the method proposed in this paper, which is available as a preprint \cite{lopez2018semantic}, with the addition of people re-identification features to guide the fusion of per-camera detections.

Among \textbf{probabilistic} methods, an interesting example is the use of a multi-view model shaped by a Bayesian network to model the relationships between occlusions \cite{peng2015robust}. Detections are here assumed to be images of either pedestrians or \textit{phantoms}, the former differentiated from the latter by inference on the network.

Recent approaches are focused on \textbf{deep learning} methods. The combination of CNNs and Conditional Random Fields (CRF) can be used to explicitly model ambiguities in crowded scenes \cite{baque2017deep}. High-order CRF terms are used to model potential occlusions, providing robust pedestrian detection.  Alternatively, multi-view detection can be handled by an end-to-end deep learning method based on an occlusion-aware model for monocular pedestrian detection and a multi-view fusion architecture \cite{chavdarova2017deep}.

\subsection{Improving detection's localization} \label{Improving detection's localization}
Algorithms in all of these groups require accurate scene calibration: small calibration errors can produce inaccurate projections and back-projections which may contravene key assumptions of the methods. These errors may lead to misaligned detections, hindering their later use. To cope with this problematic, one can rely on an Height-Adaptive Projection (HAP) procedure in which a gradient descent process is used to find both the optimal pedestrian's height and location on the ground-plane by maximizing the alignment of their back-projections with foreground masks on each camera \cite{peng2015robust}.


\begin{figure*}[t!]
  \centering
  \subfloat[]{\includegraphics[width=1\linewidth,keepaspectratio]{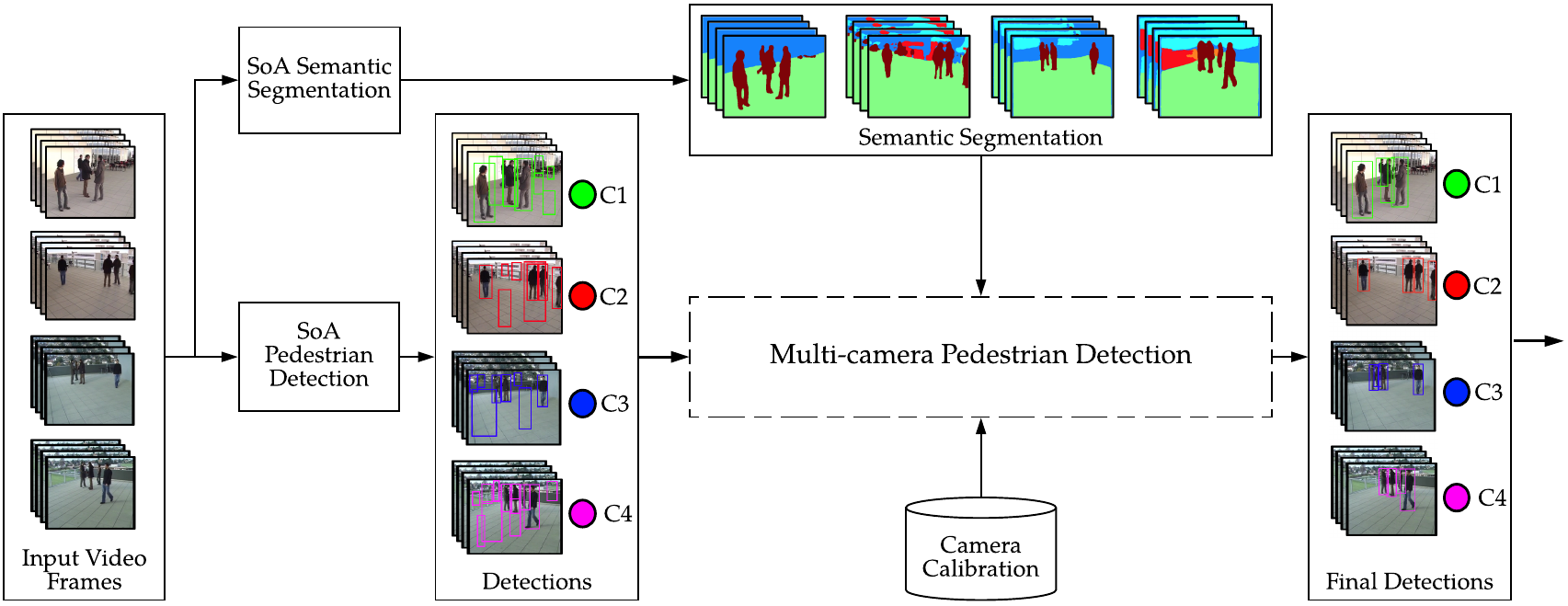}}
  \\
  \subfloat[]{\includegraphics[width=1\linewidth,keepaspectratio]{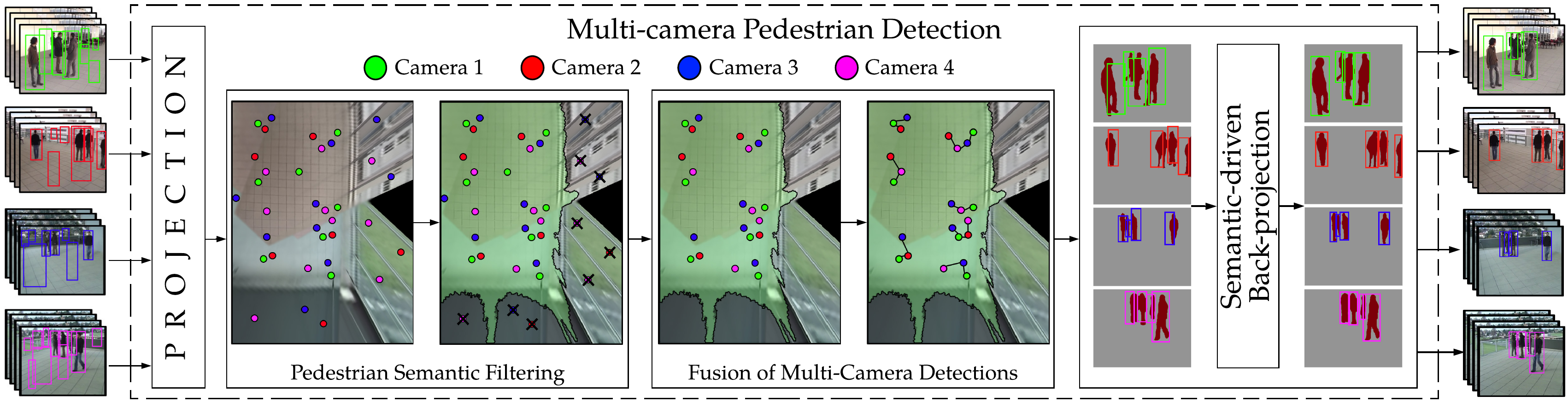}}
  \caption{Overall pedestrian detection method. Top (a): processing starts performing both a semantic segmentation and a pedestrian detection over a set of cameras (four, in the illustration) with overlapping fields of view. The segmentation, the detections and the camera calibration parameters feed the Multi-camera Pedestrian Detection module which is described in detail in bottom (b): detections are projected onto a 3D reference plane; a Pedestrian Semantic Filtering module is used to remove detections located out of the automatically generated \(\mathcal{AOI}\); the remaining detections are combined, based on a disconnected graph, to obtain global detections. The so-obtained global detections are back-projected to the camera views, and the Semantic-Driven Back-Projection module globally refines the location of these detections by also using semantic cues. Better viewed in color.}
  \label{Overall system framework}
\end{figure*}

\section{Proposed Pedestrian Detection Method} \label{Proposed Method}
The proposed method is depicted in Figure \ref{Overall system framework}. First, state-of-the-art algorithms for monocular pedestrian detection and semantic segmentation are used to extract people detections and the semantic cues for each camera respectively. These cues drive the automatic definition of the \(\mathcal{AOI}\), and detections outside this area are discarded. Surviving per-camera detections are combined to obtain global 3D detections by establishing rules and constraints on a disconnected graph. These detections are back-projected to their original camera views in order to further refine their location and height estimates.

\begin{figure*}[t!]
  \centering
  \includegraphics[width=1\textwidth,keepaspectratio]{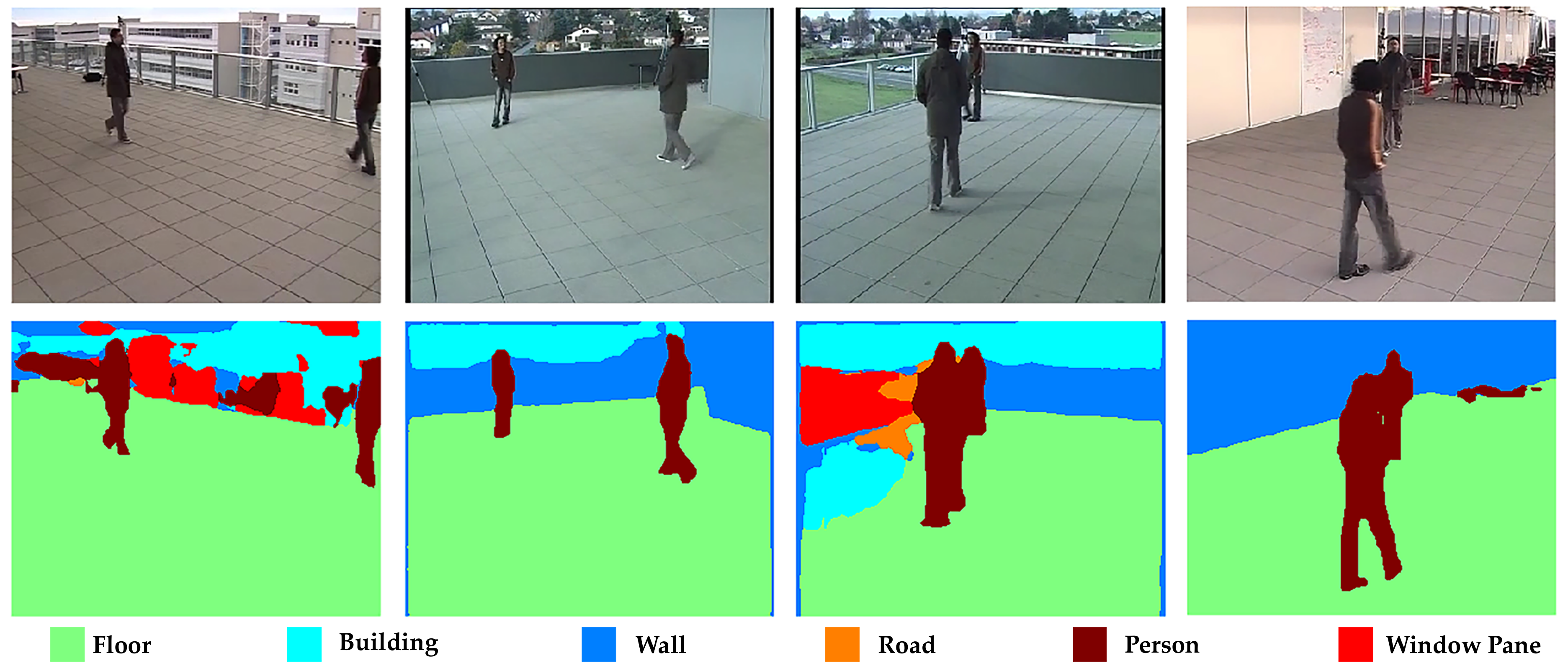}
  \caption{Top row represents RGB frames from the Terrace Dataset \cite{fleuret2008multicamera,web-Terrace}. Bottom row represents the correspondent semantic labels obtained by the PSP-Net algorithm \cite{zhao2016pyramid}. Columns from left to right represent cameras 1 to 4 of this dataset. The bottom legend indicates the detected semantic classes. Better viewed in color.}
  \label{Semantic Terrace Dataset}
\end{figure*}

\subsection{Preliminaries} \label{Projection of people detections}
\textit{Monocular Pedestrian Detection}: is performed using a state-of-the-art detector. In order to avoid a potential height-bias, we ignore the height and width of the detected bounding boxes, i.e. the \(j^{th}\) pedestrian detection at camera \(k\) is just represented by the middle point of the base of its bounding box: \(\mathbf{p}_{j,k}= (x,y,1)^T\), in homogeneous coordinates\footnote{we use common notation, upper case to denote 3D points/coordinates and lower case to denote 2D camera plane points/coordinates.}.

\textit{Semantic Segmentation}: is performed using a state-of-the-art semantic segmentation algorithm. The method is used to label each image pixel \(\mathbf{p}_k\) for every camera \(k\) and every frame \(n\): \(l_{n}(\mathbf{p}_k) = \mathit{s_i}\), where \(\mathit{s_i}\) is one of the \(L\) pre-trained semantic classes: \(S= \lbrace \mathit{s_i} \rbrace\), where \(i \in [1,L] \), i.e. \textit{floor}, \textit{building}, \textit{wall}... Figure \ref{Semantic Terrace Dataset} depicts examples of semantic labels for selected camera frames of the Terrace Dataset \cite{fleuret2008multicamera,web-Terrace}.

\textit{Projection of People Detections}: Let \(\mathcal{H}_k\) be the homography matrix that transforms points from the image plane of camera \(k\) to the world ground-plane. The \(j^{th}\) detection of camera \(k\), \(\mathbf{p}_{j,k}=(x,y,1)'\) is projected onto the ground plane by: 

\begin{equation}
\mathbf{P}_{j,k}\ = \mathcal{H}_k \times \mathbf{p}_{j,k} = (\textit{X}, \textit{Y}, \textit{T})',
\label{eq:Detection projection}
\end{equation}
which corresponds to a \((X=\textit{X}/T,Y=\textit{Y}/T,Z=0)'\) 3D point of the ground plane.

\subsection{Pedestrian Semantic Filtering} \label{subsec:Pedestrian Semantic Filtering}

\subsubsection*{Automatic Definition of the \(\mathcal{AOI}\)}
To obtain a semantic partition of the ground-plane, an adaptation of \cite{lopez2018automatic} for static-camera scenarios is carried out. We first project \textit{every} image pixel \(\mathbf{p}_k\) via \(\mathcal{H}_k\). Every projected point \(\mathbf{P}_k\) inherits the semantic label assigned to \(\mathbf{p}_k\): 
\begin{equation}
l_{n}(\mathbf{P}_k) = l_{n}(\mathbf{p}_k) = s_i \in S.
\label{eq:Semantic Projection}
\end{equation}
Thereby, a semantic locus---a ground-plane semantic partition, is obtained for each camera. The extent of each locus is defined by the image support, and missing points inside the locus are completed by nearest-neighbor interpolation.

\begin{figure*}[t!]
  \centering
  \includegraphics[width=1\linewidth,keepaspectratio]{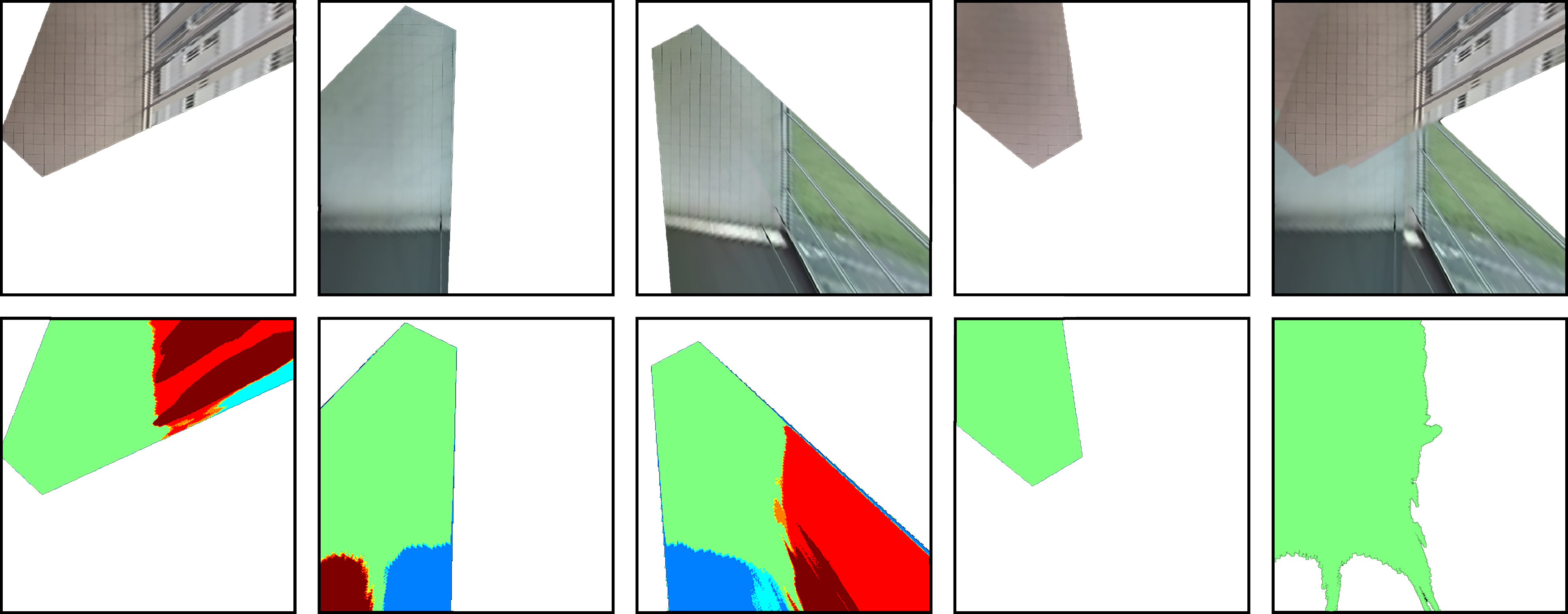}
  \caption{Temporally-smoothed projected loci for each camera (columns 1 to 4) of the Terrace Dataset \cite{fleuret2008multicamera,web-Terrace}, both in the RGB domain (top) and the semantic labels domain (bottom). The last column depicts, again in both domains, the resulting \(\mathcal{AOI}\) which, in the example, consists of the combined \textit{floor} class of the four smoothed loci. Better viewed in color.}
  \label{Bird-eye semantic views from Terrace Dataset}
\end{figure*}

In order to globally reduce the impact of moving objects and segmentation errors, we propose to temporally aggregate each locus along several frames. In a set of \(T\) loci obtained for \(T\) consecutive frames, a given point on the ground plane \(\mathbf{P}_k\) is labeled with \(T\) semantic labels, which may be different owing to inaccuracies in the semantic segmentation or to the presence of moving objects. A single temporally-smoothed label \(\bar{l}_n(\mathbf{P}_k)\) is obtained as the mode value of this set. Examples of these per-camera obtained smoothed loci are included in the first four-columns of Figure \ref{Bird-eye semantic views from Terrace Dataset}.

We propose to combine these loci to define the \(\mathcal{AOI}\). The definition of the \(\mathcal{AOI}\) is scenario-dependent but can be generalized by defining a set \(\mathcal{G}\) of ground-related semantic classes: \textit{floor}, \textit{grass}, \textit{pavement}, etc. The operational area \(\mathcal{AOI}\) is obtained as the union of the projected pixels from any camera which are labeled with any class in \(\mathcal{G}\):

\begin{equation}
\mathcal{AOI} = \bigcup_k^K \mathbf{P}_k,\quad s.t.\quad \bar{l}_n(\mathbf{P}_k) \in \mathcal{G}.
\label{eq:AOI}
\end{equation}

An example of a so-obtained \(\mathcal{AOI}\) is included in the right-most column of Figure \ref{Bird-eye semantic views from Terrace Dataset}.

\subsubsection*{Detection Filtering}
Projected detections \(\mathbf{P}_{j,k}\) lying outside the operational area, \(\mathbf{P}_{j,k} \notin \mathcal{AOI}\), are filtered out and so, discarded for forthcoming stages.

\subsection{Fusion of Multi-Camera Detections} \label{subsec:Multi-camera Detection}
We propose a \textbf{geometrical} approach to combine detections on the ground-plane. Every camera single detection is considered a vertex of a disconnected graph located in the reference plane. Vertices are then joined generating connected components \(C_m\), each representing a joint 3D global detection. The whole fusion process is summarized in Figure \ref{fig:GraphAlgorithm}. The conditions that shall be satisfied to join two vertices or detections, \(\mathbf{P}_{j,k}\) and \(\mathbf{P}_{j',k'}\), are:

\begin{figure}[t!]
    \centering
  \includegraphics[width=\columnwidth,keepaspectratio]{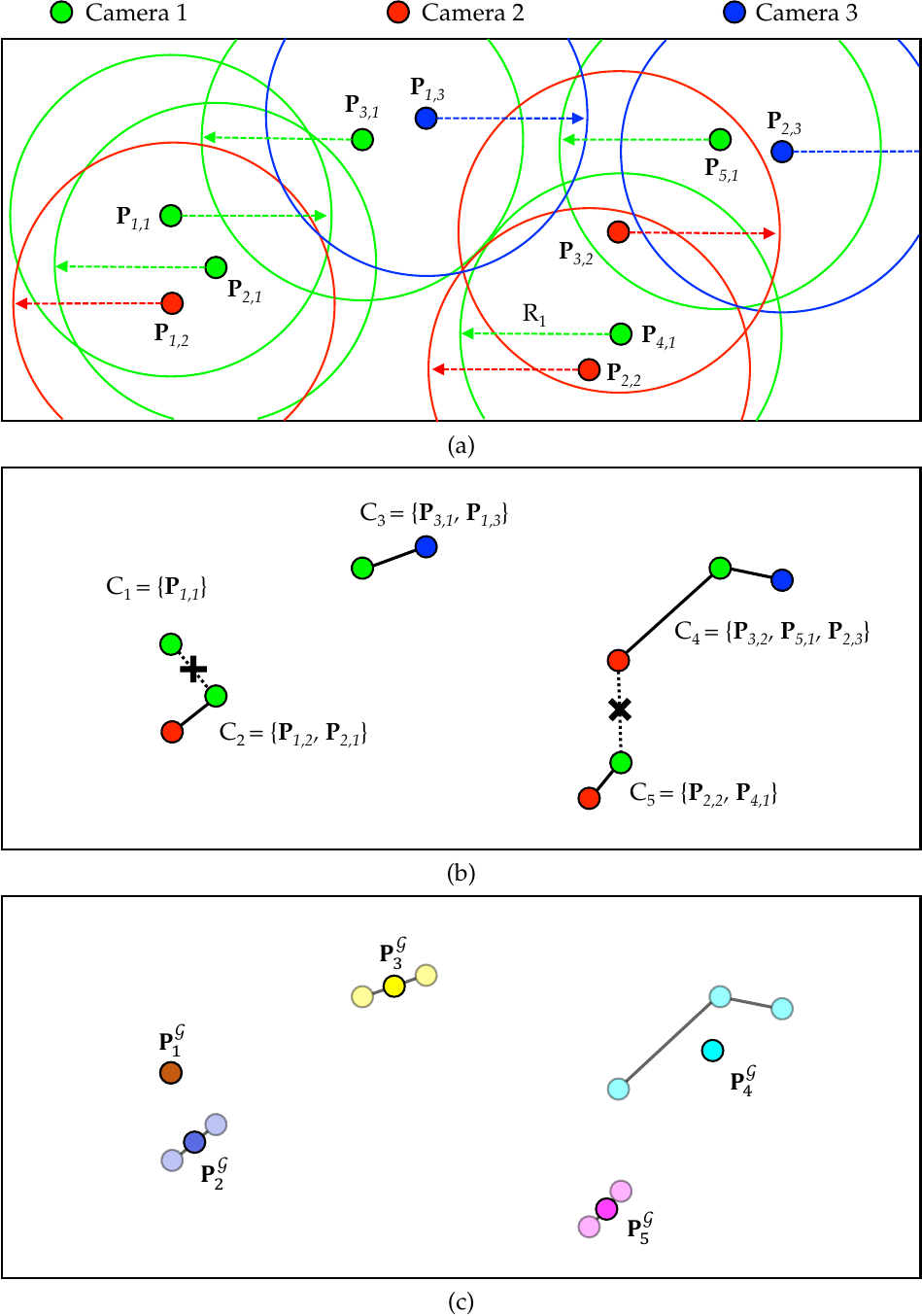}
    \caption{Fusion of multi-camera detections in the ground plane. (a) The distance \(R_1\), depicted here as circumferences around detections, defines neighbors for each detection \(\mathbf{P}_{j,k}\). (b) Connected components \(C_m\) are defined for detections: (i) which \(l_2\)-\(norm\) is lower than \(R_1\) and (ii) that are projected from different cameras. Connected components fulfilling (i) but not (ii) are represented by dashed lines crossed out. (c) The ground-plane detection \(\mathbf{P}^\mathcal{G}_m\) is obtained as the arithmetic mean of all the detections in a connected component \(C_m\). Better viewed in color.}
    \label{fig:GraphAlgorithm}
\end{figure}

\begin{enumerate}
  \item That vertices in a connected component are \textit{close enough}. The \(l_2\)-\(norm\) between any two vertices in \(C_m\) shall be smaller than a predefined distance \(R_1\): \( \lVert \mathbf{P}_{j,k}, \mathbf{P}_{j',k'} \rVert _{2} \leq R_1\) (Figure \ref{fig:GraphAlgorithm} (a)). \(R_1\) may be fixed in the interval between \(2.5\) and \(3.5\) with no influence in the results. We experimentally set \(R_1 = 3\) meters to: 1) reduce the computational cost of the final stage (see below) assuming that vertices separated \(R_1\) do not belong to the same object and 2) protect against calibration errors, assuming that they are not larger than \(R_1\).
  \item That vertices in a connected component come from different cameras. This condition prevents the joining of two different detections from the same camera which are near in the ground plane. (Figure \ref{fig:GraphAlgorithm} (b))
\end{enumerate}
To avoid ambiguities, the creation of connected components is performed in order, according to the spatial position of the detections: those with a lower module are combined first.

The outcome of the fusion process for \(K\) cameras is a set of \(M\) connected components \(\lbrace C_m,\quad m= 1,...,M \rbrace\), each containing \(K_m\) detections: \(\mid C_m \mid = K_m \leq K\), where \(K_m < K\) when a person is occluded or not detected in one or more cameras.

As each connected component is assumed to represent a single person, an initial ground-position of the person \({\mathbf{P}}_m^\mathcal{G} = (X_m, Y_m, Z_m=0)^T \) is obtained by simply computing the arithmetic mean of all the detections in the \(C_m\) connected components (Figure \ref{fig:GraphAlgorithm} (c)).

\subsection{Semantic-Driven Back-Projection} \label{subsec:Semantic Adaptive Back-Projection (SABP)}
To obtain correctly positioned detections, i.e. visually precise detections, in each camera, ground-plane detections need to be back-projected to each camera and 2D bounding-boxes enclosing pedestrians need to be outlined based on these projections.

\begin{figure}[t!]
    \centering
    \includegraphics[width=1\columnwidth,keepaspectratio]{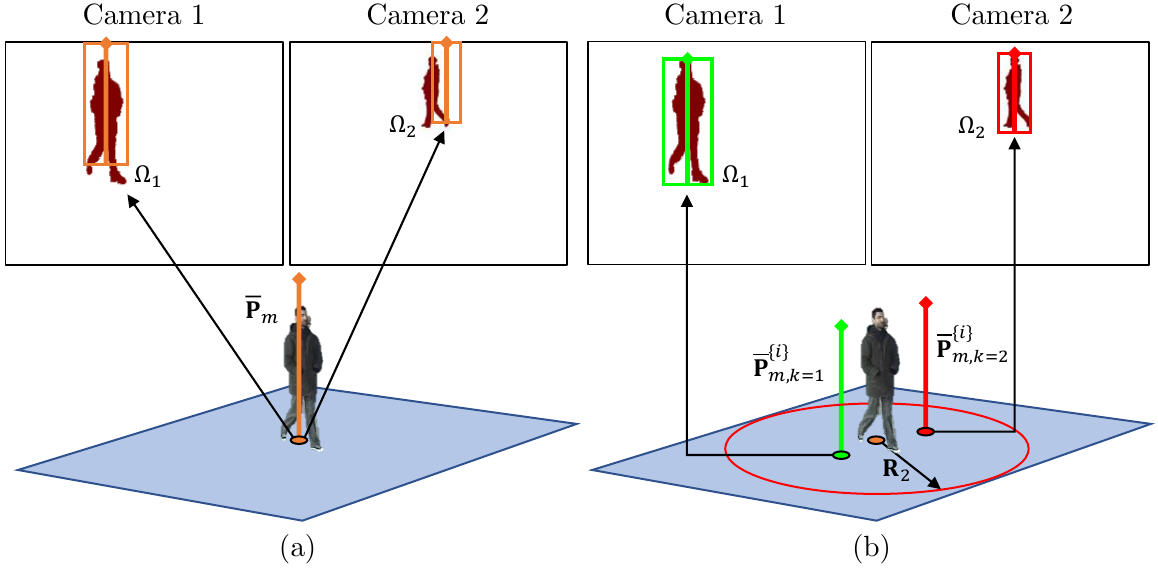}
    \caption{(a) Back-projecting global segment \(\overline{\mathbf{P}}_{m}\) results in misaligned bounding-boxes due to pedestrian self-occlusion, calibration errors and the uncertainty on the pedestrians' height. (b) The proposed optimization process results in the best-aligned segments \(\overline{\mathbf{P}}_{m,k}\) for each camera. Better viewed in color.}
    \label{fig:OptimizationAlgorithm}
\end{figure}

\subsubsection*{The problem of back-projecting 3D detections}
Let \(\overline{\mathbf{P}}_{m}\) be an orthogonal line segment to the ground plane which represents the detected pedestrian and extends from the detection \({\mathbf{P}}_m^\mathcal{G}\) to a 3D point \(h_m\) meters above. Using the camera calibration parameters, the segment \(\overline{\mathbf{P}}_{m}\) can be back-projected onto camera \(k\). This back-projection defines a 2D line segment \(\overline{\mathbf{p}}_{m,k}\), which extends between \(\mathbf{p}_{m,k}\) and \(\mathbf{p}_{m,k} + \vec{\eta}\) (see Figure \ref{fig:OptimizationAlgorithm} (a)).

We propose to create 2D bounding-boxes around these back-projected 2D line segments. To this aim, each segment is used as the vertical middle axis of its associated 2D bounding-box \(\mathbf{b}_{m,k}\). For simplicity, the width of \(\mathbf{b}_{m,k}\) is made proportional to its height. Due to pedestrian self-occlusion, calibration errors and the uncertainty on the pedestrians' height, this back-projection process results in misaligned bounding-boxes (see Figure \ref{fig:OptimizationAlgorithm} (a)), hindering their later use and degrading camera-wise performance. 

To handle this problematic, we define an iterative method which aims to globally optimize the alignment between all 3D detections and their respective views or back-projections in all cameras. This method is based on the idea proposed in \cite{peng2015robust}. While the referenced method is guided by a foreground-segmentation, we instead propose to use a cost-function driven by the set of pedestrian-labeled pixels \(\Omega_k\) from the semantic segmentation (e.g. see person label in Figure \ref{Semantic Terrace Dataset}). Next, we detail the full process for the sake of reproducibility.

\subsubsection*{Method Overview}
As a 3D detection \(\overline{\mathbf{P}}_{m}\), with height \(h_m\), inevitably results in misaligned back-projected 2D detections, the proposed method tries to adapt the 3D detection segment to each camera, generating a set of 3D detection segments, \(\overline{\mathbf{P}}_{m,k}\), for each 3D detection and iteratively modifying their positions and height to maximize 2D detections' alignment with the semantic segmentation masks, while constraining all the segments to have the same final height \(h'_{m}\) (as they are all projections of a same pedestrian) and to be located sufficiently close to each other. This process is not performed independently for each 3D detection but jointly and iteratively for all 3D detections. Observe that the joint nature of the optimization problem for all 3D detections is a key step as pedestrian pixels \(\Omega_k\) may contain segmentations from more than one pedestrian.

For each 3D segment \(\overline{\mathbf{P}}_{m}\), the method starts by initializing (i.e., iteration \(i=0\)) the per-camera adapted segments:

\begin{equation}
  \overline{\mathbf{P}}^{\left(  i=0 \right)}_{m,k} = \overline{\mathbf{P}}_m \:\:,\: k=1...K.
  \label{eq:1}
\end{equation}

\subsubsection*{Iterative steepest-ascent algorithm}
For each 3D segment, let \(\mathcal{P}^{\left( i \right)}_k = \lbrace \overline{\mathbf{P}}^{\left( i\right)}_{m,k}, \, m=1...M \rbrace\) be the set of adapted detections to camera \(k\) at iteration \(i\), and let \(\mathbb{P}^{\left( i\right)}= \lbrace \mathcal{P}^{\left( i\right)}_k, \, k=1...K \rbrace \) be the set of camera-adapted segments for all cameras at the same iteration.

The optimization process aims to find \(\mathbb{P^*}\), the solution to the constrained optimization problem:

\begin{equation}
\begin{split}
  \mathbb{P^*} & = {arg\,max}_{\mathbb{P}}\,\Psi(\mathbb{P}),\enspace s.t \enspace \lVert \overline{\mathbf{P}}_{m}, \overline{\mathbf{P}}_{m,k} \rVert_{2} \leq R_2 \, \forall (m,\,k),
  \end{split}
  \label{eq:Optimization problem}
\end{equation}
where \(R_2\) defines the maximum distance between 3D projections of a single pedestrian, which we set to twice the average width of the human body, i.e. 1 meter, to forestall the effect of nearby pedestrians in the image plane. Performed experiments suggest that variations in \(R_2\) value have no significant influence on the results. 

\(\Psi(\mathbb{P})\) is defined as the cost function to maximize and is based on the alignment of the back-projected bounding boxes with the set of pedestrian-labeled pixels in each camera: \(\Omega_k\). The cost function considers the information from all the cameras. 

\begin{equation}
    \Psi(\mathbb{P}^{\left( i\right)})=-\sum_{k=1}^K \frac{\sum_{\mathbf{p}} \,  \gamma(\mathbf{p},\Omega_{k}) \, \Phi(\mathbf{p},\mathbb{P}^{\left( i\right)})}{|F_{k}|},
    \label{eq:H function}
\end{equation}
where \(\gamma(\mathbf{p},\Omega_{k})\) is a weight for pixel \(\mathbf{p}\): \(\omega\) for pedestrian and \(\omega/3\) for non pedestrian pixels, \(\omega=1\) in our setup, \(|F_{k}|\) is the number of pixels in the camera image plane and \(\Phi(\mathbf{p},\mathbb{P})\) is the loss function of pixel \(\mathbf{p}\) with respect to \(\mathbb{P}\):

\begin{equation}
  \Phi(\mathbf{p},\mathbb{P}^{\lbrace i \rbrace}) = 
    \begin{cases} 
          \prod_{m|\mathbf{p} \in \mathbf{b}^{\left( i\right)}_{m,k}} (1-1/d_{m,k}))      & \mathrm{, \, if} \, l_{k}(\mathbf{p}) \in \Omega_{k}
          \\ 
          \\
          1 - \prod_{m|\mathbf{p} \in \mathbf{b}^{\left( i\right)}_{m,k}} (1-1/d_{m,k}))  & \mathrm{, \, if} \, l_{k}(\mathbf{p})  \notin \Omega_{k},
    \end{cases}
  \label{eq:LH function}
\end{equation}
where \(d_{m,k}\) is the distance from \(\mathbf{p}\) to the vertical middle axis \(\overline{\mathbf{p}}^{\left( i\right)}_{m,k}\) of the back-projected bounding box \(\mathbf{b}^{\left( i\right)}_{m,k}\).

At each iteration \(i\), the set of camera-adapted segments is moved towards the direction of maximum increment:

\begin{equation}
  \mathbb{P}^{\left( i\right)} = \mathbb{P}^{\left( i-1\right)} + \tau_{i} \overrightarrow{\nabla}\Psi(\mathbb{P}^{\left( i-1\right)}),
  \label{eq:2}
\end{equation}
where \(\tau_i \in \mathbb{R}_{+}\) is the gradient-step that makes \(\mathbb{P}^{\left( i\right)} \geq \mathbb{P}^{\left( i-1\right)}\). This gradient-step is initialized, \(\tau_{0} = 5\), and updated following a decrease schedule of \(50\%\) every 3 iterations, to ease convergence. The gradient \(\overrightarrow{\nabla}\Psi(\mathbb{P}^{\left( i\right)})\) in the \(i\)-th iteration is approximated by Forward Difference Approximation:
\begin{equation}
  \overrightarrow{\nabla}\Psi(\mathbb{P}^{\left( i\right)}) = \frac{ \Psi(\mathbb{P}^{\left( i\right)}) - \Psi(\mathbb{P}^{\left( i\right)} - \epsilon)}{\epsilon}.
  \label{eq:Finite Central Difference}
\end{equation}

The algorithm continues until convergence is reached or the \(R_2\)-constrain is violated.


\section{Results} \label{Results}
This section addresses the evaluation of the proposed method. To this aim, we first describe the evaluation framework; then, in the ablation studies, we measure the performance improvement of each of the method's stages; finally, we finish by comparing our approach with alternative state-of-the-art approaches in classic and recent multi-camera datasets.

\subsection{Evaluation Framework}

\subsubsection*{Datasets}
Results are obtained by evaluating the proposed method over five scenarios extracted from four publicly available multi-camera datasets in which cameras are calibrated and temporally synchronized:

\begin{itemize}
\item EPFL Terrace \cite{fleuret2008multicamera,web-Terrace}: Generally used in the state-of-the-art to evaluate multi-camera approaches. It consists of a 5000 frames sequence per camera showing up to eight people walking on a terrace captured by four different cameras. All the cameras record a close-up view of the scene.
\item EPFL RLC \cite{baque2017deep,web-RLC}: Consists of an indoor sequence of 2000 frames per camera recorded in the EPFL Rolex Learning Center using three static HD cameras with overlapping field of views. All these cameras represent close-up views of the scene.
\item EPFL Wildtrack \cite{chavdarova-et-al-2018,web-wildtrackDataset}: A challenging multi-camera dataset which has been explicitly designed to evaluate deep learning approaches. It has been recorded with 7 HD cameras with overlapping fields of view. Pedestrian annotations for 400 frames are provided. All of them are used to define the evaluation set used in this paper.
\item PETS 2009 \cite{web-PETS}: The most used video sequences from this widely used benchmark dataset have been chosen.
\begin{itemize}
\item PETS 2009 S2 L1, which contains 795 frames recorded by eight different cameras of a medium density crowd---in this evaluation, we have just selected 4 of these cameras: view 1 (far field view) and views 5, 6 and 8 (close-up views). 
\item PETS 2009 City Center (CC), recorded only using two far-field view cameras with around 1 minute of annotated recording (400 frames per camera).
\end{itemize}
\end{itemize}

\begin{table*}
    \renewcommand{\arraystretch}{2}
    \begin{centering}
    \resizebox{\textwidth}{!}{
    \begin{tabular}{lccccccc}
        \cline{2-8} 
         & Resolution & \makecell{Cameras} & \makecell{Frames} & \makecell{Pedestrian \\ Density} & \makecell{Occlusions \\ Level} & \makecell{Camera \\ PoV} & \makecell{GT Annotations \\ (Cameras)}\tabularnewline
        \hline 
        PETS 2012 CC \cite{web-PETS} & 768$\times$576 & 2 & 795 & + & + & High & Frame (1)\tabularnewline
        PETS 2012 S2 L1 \cite{web-PETS} & 768$\times$576 & 4 & 795 & + & + & High & Frame (1)\tabularnewline
        EPFL Terrace \cite{fleuret2008multicamera,web-Terrace} & 366$\times$288 & 4 & 5008 & ++ & ++ & Low & Frame (4)\tabularnewline
        EPFL RLC \cite{baque2017deep,web-RLC} & 480$\times$270 & 3 & 1197 & ++ & ++ & Low & Frame (1)\tabularnewline
        EPFL Wildtrack \cite{chavdarova-et-al-2018,web-wildtrackDataset} & 1920$\times$1080 & 7 & 401 & +++ & +++ & Medium & Ground Plane (NA)\tabularnewline
        \hline 
    \end{tabular}}
    \caption{Dataset description. Pedestrian Density and Occlusions Level are ranked subjectively in terms of the number of pedestrians with respect to the scenario space and the number and detection difficulty of the occlusions (\(+\) stands for easy, \(++\) stands for medium and \(+++\) stands for difficult). The Camera PoV (Point of View), which affects the level of pedestrians' occlusions, is categorized as Low, Medium or High according to the height of the camera with respect to the ground-plane. Figure \ref{fig:QualitativeResults} depicts frame examples for these analyzed datasets.}
    \label{tab:Dataset Comparison}
    \par\end{centering}
\end{table*}

Table \ref{tab:Dataset Comparison} contains a comparative description of these datasets including the type of data and annotations provided, as well as a subjective indication of their complexity for the pedestrian detection task.

\subsubsection*{Performance Indicators}
To obtain quantitative performance statistics according to an experiment-based evaluation criterion the following state-of-the-art performance indicators have been selected: Precision (P), Recall (R), F-Score (F-S), Area Under the Curve (AUC), N-MODA (N-A) and N-MODP (N-P) \cite{dollar2009pedestrian,stiefelhagen2006clear}. To globally assess performance, a single value for each statistic and each configuration is provided by averaging per-camera ones.

\subsection{System Setup}
A common setup has been used for all the presented results. Faster-RCNN \cite{renNIPS15fasterrcnn}, YOLOv3 \cite{redmon2018yolov3} and EfficientDet-D7 \cite{tan2020efficientdet} are used as baseline algorithms to obtain monocular pedestrian detections. The three object detectors are pre-trained on the COCO dataset \cite{lin2014microsoft} and we do not fine-tune nor adapt them to any of the faced scenarios. For the semantic segmentation, the Pyramid Scene Parsing Network (PSP-Net) \cite{zhao2016pyramid}, pre-trained on the ADE20K dataset \cite{zhou2017scene} (\(L=150\), has been selected considering a trade-off between performance and efficiency.)

In the Pedestrian Semantic Filtering stage, all frames in each sequence are used for temporal and spatial semantic aggregation, i.e. \(T=N\). For the Semantic-Driven Back-Projection stage, the initial height estimation \(h_{m}\) has been set to an average pedestrian height of \(1.7\)m. Besides, for all the datasets, convergence in the iterative steepest-ascent algorithm has been reached before or at the \(8^{th}\) iteration.

\subsection{Results Overview}
The evaluation has been performed carrying out two different studies:

\begin{itemize}
\item The Ablation Studies aim to gauge the impact of the different stages in the performance of the proposed approach. To this end, the following versions of the proposed method are compared:
\begin{enumerate}
    \item ``Baseline (Faster-RCNN, YOLOv3 and EfficientDet-D7)'', provides reference results of monocamera pedestrian detectors.
    \item ``Baseline + Filtering (\textit{Filt})'' is a simplified version of our method which aims to independently evaluate the effect of the proposed automatic \(\mathcal{AOI}\) computation obtained by the ``Pedestrian Semantic Filtering" stage.
    \item ``Baseline + Filtering (\textit{Filt}) + Fusion (\textit{Fus}) + Back-Projection (\textit{BP})'' is the full version of the proposed method, which additionally evaluates the ``Fusion of Multi-Camera Detections" and ``Semantic-Driven Back-Projection" stages.
\end{enumerate}
Ablation Studies are conducted on four of the described datasets: Terrace, PETS 2009 S2 L1, PETS 2009 CC and RLC.

\item State-of-the-art Comparison results analyze the proposed method with respect to several non deep-learning state-of-the-art multi-camera pedestrian detectors on the same four scenarios used in the Ablation Studies. Additionally, the method is compared with novel deep-learning methods on the Wildtrack dataset.
\end{itemize}

\subsection{Ablation Studies}
\subsubsection*{Evaluation Criterion}\label{Ex1: Evaluation Criteria}
The availability of bounding-box annotations permits to use the classic performance criterion \cite{garcia2015people}: a detection is considered a TP one if the Intersection Over Union (IoU) with a ground-truth bounding-box is higher than \(0.5\).

\subsubsection*{Results}
\begin{figure*}[t!]
    \centering
    \includegraphics[width=0.75\textwidth,keepaspectratio]{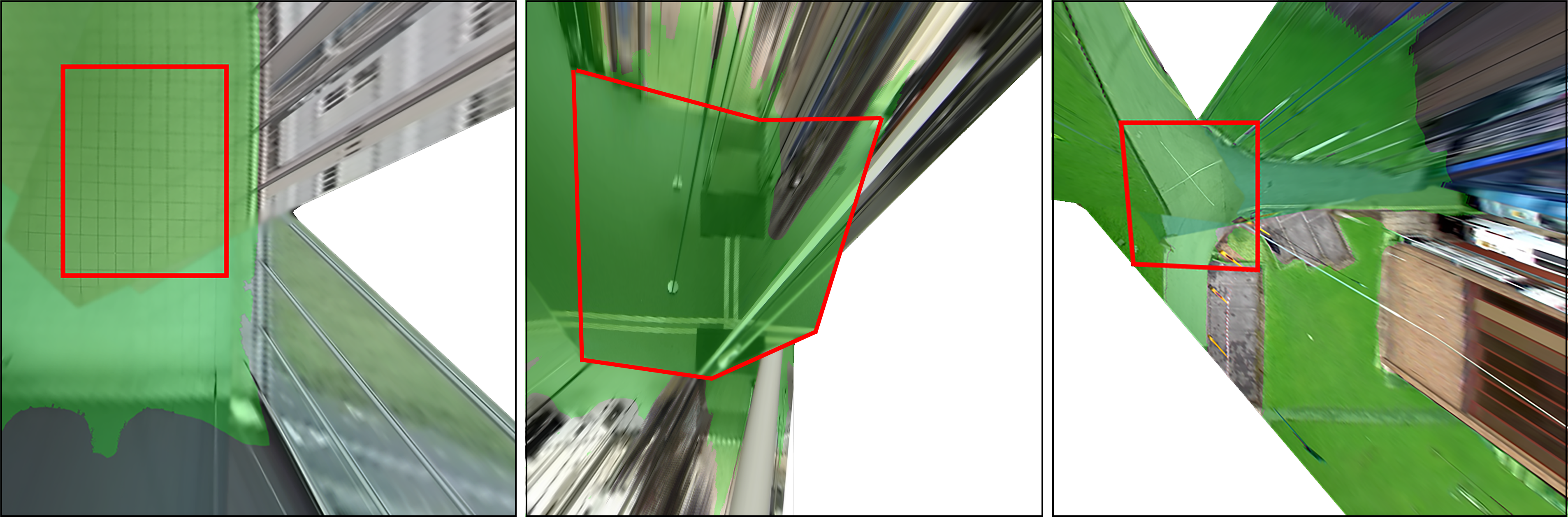}
    \caption{Ablation Studies: Automatically obtained \(\mathcal{AOI}\) (superimposed in green) compared to the \(\mathcal{AOI}\) manually annotated (red box) by the authors of EPFL Terrace \cite{fleuret2008multicamera,web-Terrace} (left), EPFL RLC Dataset \cite{baque2017deep,web-RLC} (middle) and PETS2009 \cite{web-PETS} (right). Better viewed in color.}
    \label{fig:AOI Results}
\end{figure*}

Table \ref{tab:Ablation Results} agglutinates the method's performance on a per-stage basis. Qualitative examples of automatically generated \(\mathcal{AOI}\)s and algorithm results are depicted in Figure \ref{fig:AOI Results} and Figure \ref{fig:QualitativeResults} respectively. A visual example of the limitations of the Semantic-Driven Back-Projection stage is included in Figure \ref{fig:Problems Optimization}. 


\begin{table*}[t!]
\begin{centering}
    \scriptsize
    \centerline{%
    \renewcommand{\arraystretch}{2}
    \resizebox{\textwidth}{!}{
    \begin{tabular}{c c c c c c c c c c c c c c c c c c c}
	\cline{4-19} 
	\multicolumn{1}{c}{} & \multicolumn{1}{c}{} &  & \multicolumn{16}{c}{Dataset}\tabularnewline
	\cline{4-19} 
	\multicolumn{1}{c}{} & \multicolumn{1}{c}{} &                            &                 \multicolumn{4}{c}{PETS 2009 CC}              &              \multicolumn{4}{c}{PETS 2009 S2 L1}              &          \multicolumn{4}{c}{EPFL Terrace}                     &                     \multicolumn{4}{c}{EPFL RLC}              \tabularnewline
	\cline{2-19}                                                                                                                             
	\multicolumn{1}{c}{}                           & Filt       & \makecell{Fus \& \\ BP}  & AUC           & F-S           & NA            & NP            & AUC           & F-S           & NA           & NP             & AUC           & F-S           & NA            & NP            & AUC           & F-S           & NA            & NP            \tabularnewline
	\hline                                                                                                                                   
	\multirow{3}{*}{\rotatebox{90}{Faster-RCNN}}   &            &            & 0.90          & 0.91          & 0.85          & 0.76          & 0.90          & 0.91          & 0.85          & 0.76          & 0.82          & 0.84          & 0.71          & 0.74          & 0.77          & 0.78          & 0.58          & 0.69          \tabularnewline
	                                               & \checkmark &            & 0.90          & 0.91          & 0.85          & 0.76          & 0.90          & 0.91          & 0.85          & 0.76          & 0.84          & 0.85          & 0.73          & 0.74          & 0.80          & 0.82          & 0.68          & 0.70          \tabularnewline
	                                               & \checkmark & \checkmark & \textbf{0.94} & \textbf{0.94} & \textbf{0.88} & \textbf{0.79} & \textbf{0.92} & \textbf{0.93} & \textbf{0.89} & \textbf{0.79} & \textbf{0.87} & \textbf{0.90} & \textbf{0.83} & \textbf{0.77} & \textbf{0.81} & \textbf{0.81} & \textbf{0.68} & \textbf{0.70} \tabularnewline
	\hline                                                                                                                                   
	\multirow{3}{*}{\rotatebox{90}{YOLOv3}}        &            &            & 0.92          & 0.92          & 0.87          & 0.79          & \textbf{0.96} & \textbf{0.96} & \textbf{0.92} & \textbf{0.67} & 0.83          & 0.87          & 0.76          & 0.73          & 0.80          & 0.78          & 0.59          & 0.72          \tabularnewline
	                                               & \checkmark &            & 0.92          & 0.92          & 0.87          & 0.79          & 0.96          & 0.96          & 0.92          & 0.67          & 0.84          & 0.87          & 0.77          & 0.73          & 0.85          & 0.83          & 0.66          & 0.72          \tabularnewline
	                                               & \checkmark & \checkmark & \textbf{0.94} & \textbf{0.94} & \textbf{0.88} & \textbf{0.79} & 0.93          & 0.92          & 0.89          & 0.67          & \textbf{0.86} & \textbf{0.89} & \textbf{0.85} & \textbf{0.76} & \textbf{0.85} & \textbf{0.83} & \textbf{0.68} & \textbf{0.72} \tabularnewline
	\hline                                                                                                                                   
	\multirow{3}{*}{\rotatebox{90}{Efficient Det}} &            &            & \textbf{0.97} & \textbf{0.97} & \textbf{0.94} & \textbf{0.66} & \textbf{0.97} & \textbf{0.97} & \textbf{0.94} & \textbf{0.67} & 0.82          & 0.84          & 0.71          & 0.78          & 0.82          & 0.80          & 0.61          & 0.72          \tabularnewline
	                                               & \checkmark &            & 0.97          & 0.97          & 0.94          & 0.66          & 0.97          & 0.97          & 0.94          & 0.66          & 0.82          & 0.87          & 0.76          & 0.78          & 0.85          & 0.83          & 0.68          & 0.72          \tabularnewline
	                                               & \checkmark & \checkmark & 0.95          & 0.94          & 0.88          & 0.75          & 0.94          & 0.94          & 0.88          & 0.67          & \textbf{0.86} & \textbf{0.89} & \textbf{0.83} & \textbf{0.76} & \textbf{0.83} & \textbf{0.84} & \textbf{0.71} & \textbf{0.72} \tabularnewline
	\hline 
    \end{tabular}}}
    \caption{Ablation Studies: Stage-wise performance of the proposed method when Faster-RCNNN \cite{renNIPS15fasterrcnn}, YOLOv3 \cite{redmon2018yolov3} and EfficientDet \cite{tan2020efficientdet} are used as baselines. Bold values indicate best result in terms of N-MODA. Indicators are Area Under the Curve (AUC), F-Score (F-S), N-MODA (N-A) and N-MODP (N-P). \textit{Filt} stands for the ``Pedestrian Semantic Filtering" stage and \textit{Fus} \& \textit{BP} stands for the ``Fusion of Multi-Camera Detections (\textit{Fus}) and Semantic-Driven Back-Projection (\textit{BP})" stages. Datasets are sorted from left to right in terms of complexity according to Table \ref{tab:Dataset Comparison}.}
    \label{tab:Ablation Results}
    \par\end{centering}
\end{table*}

\subsubsection*{Discussion}
Table \ref{tab:Ablation Results} shows that filtering-out detections using automatically generated \(\mathcal{AOI}\)s (Baseline + Filtering) improves the performance of all the baselines for datasets where the ground-plane area does not cover the whole image representation, i.e. datasets containing close-up views of the scene as EPFL Terrace and RLC. In these datasets, our precise \(\mathcal{AOI}\)s reduce \textit{phantom} detections obtained by the baseline detectors. Although \(\mathcal{AOI}\)s are automatically computed, they are more precise (tighter to real scene edges) than those defined in the dataset. 

Overall, in the EPFL Terrace dataset, the performance of Faster-RCNN + Filtering improves Faster-RCNN by \(2.44\%\) and \(2.82\%\) in terms of AUC and N-MODA respectively. YOLOv3 + Filtering presents relative increments over YOLOv3 baseline of \(1.20\%\) and \(1.31\%\) for AUC and N-MODA respectively. Finally, EfficientDet + Filtering also overcomes its baseline results by a \(7.04\%\) for N-MODA.

For the EPFL RLC dataset with the proposed \(\mathcal{AOI}\), Faster-RCNN is improved by a \(5.13\%\) regarding AUC and by a \(17.24\%\) concerning N-MODA. For YOLOv3, relative increments of a \(6.25\%\) and a \(11.86\%\) in terms of AUC and N-MODA are achieved. EfficientDet gains relative increments of a \(3.65\%\) and a \(11.47\%\) for AUC and N-MODA.

\begin{figure*}[t!]
    \centering
    \includegraphics[width=1\textwidth,keepaspectratio]{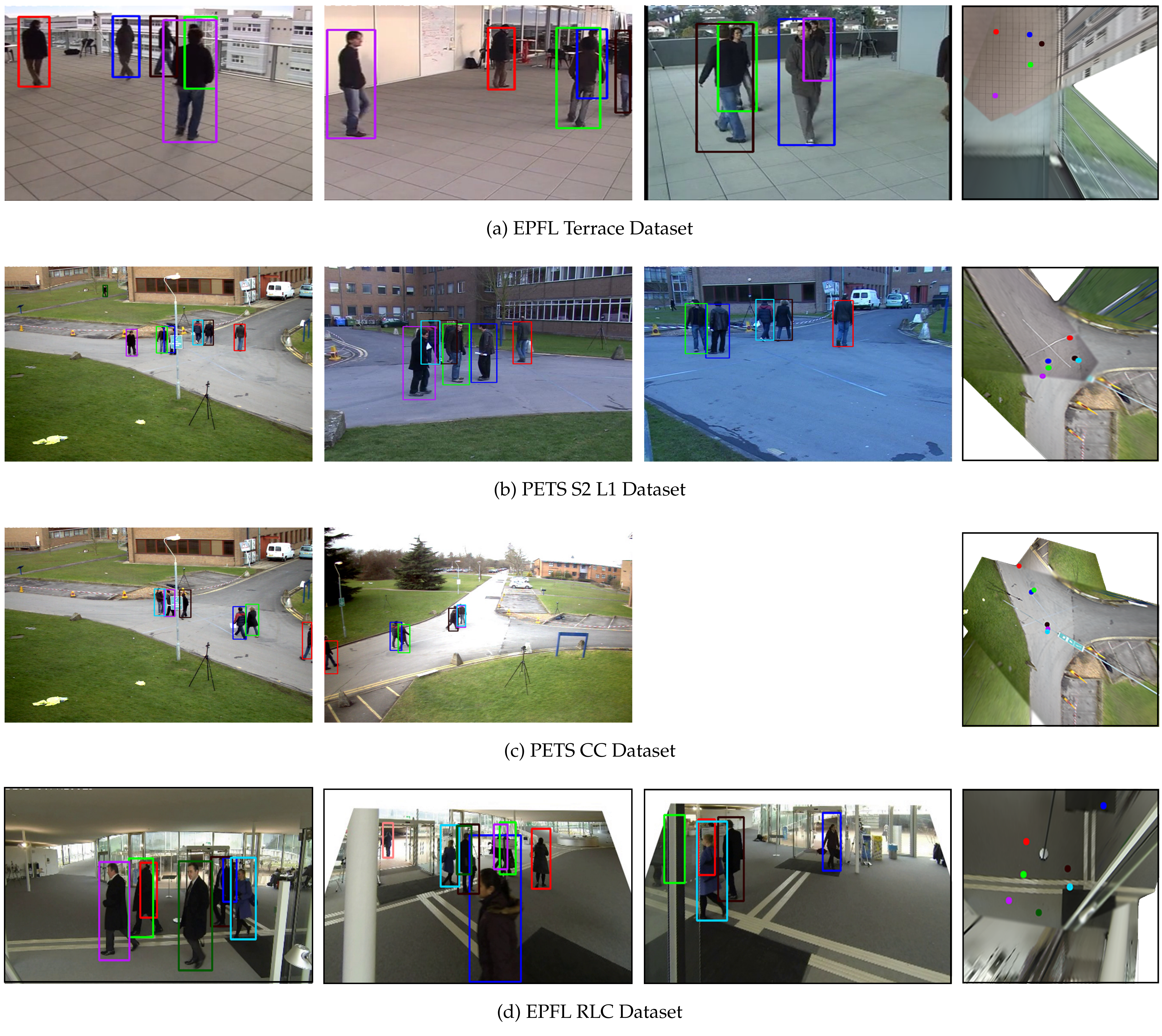}
    \caption{Ablation Studies: Proposed method qualitative results on selected frames of the EPFL Terrace, PETS S2 L1, PETS CC and EPFL RLC datasets (Faster-RCNN baseline is here used). From left to right: First three columns depict a same time frame captured by three available cameras, showing color bounding boxes (a color per pedestrian) corresponding to the final per-camera detections. The most-right column depicts obtained detections---one per pedestrian in the scene---on the ground plane, conserving the identifying colors. Better viewed in color.}
    \label{fig:QualitativeResults}
\end{figure*}

The proposed Filtering stage does not improve baselines' performance for those datasets in which the ground-plane dominates the scene, i.e. those recorded with far-field view cameras as both scenarios from PETS 2009. In these cases, although the baseline pedestrian detectors may create \textit{phantom} detections, those lie inside the proposed \(\mathcal{AOI}\) and no false-pedestrians are suppressed. However, as depicted in Figure \ref{fig:AOI Results}, the automatically obtained \(\mathcal{AOI}\)s are larger and more precise than the original operational areas in the datasets, thereby obtaining a more realistic and exhaustive evaluation. Furthermore, observe how the proposed generation method effectively handles multi-class ground partitions as in the PETS 2009 dataset, where the proposed \(\mathcal{AOI}\) encompasses \textit{road}, \textit{grass}, \textit{pavement} and \textit{side-walks} classes enabling a high adaptability to unseen scenarios (see Figure \ref{fig:AOI Results} right).

\begin{figure*}[t!]
    \centering
    \includegraphics[width=1\textwidth,keepaspectratio]{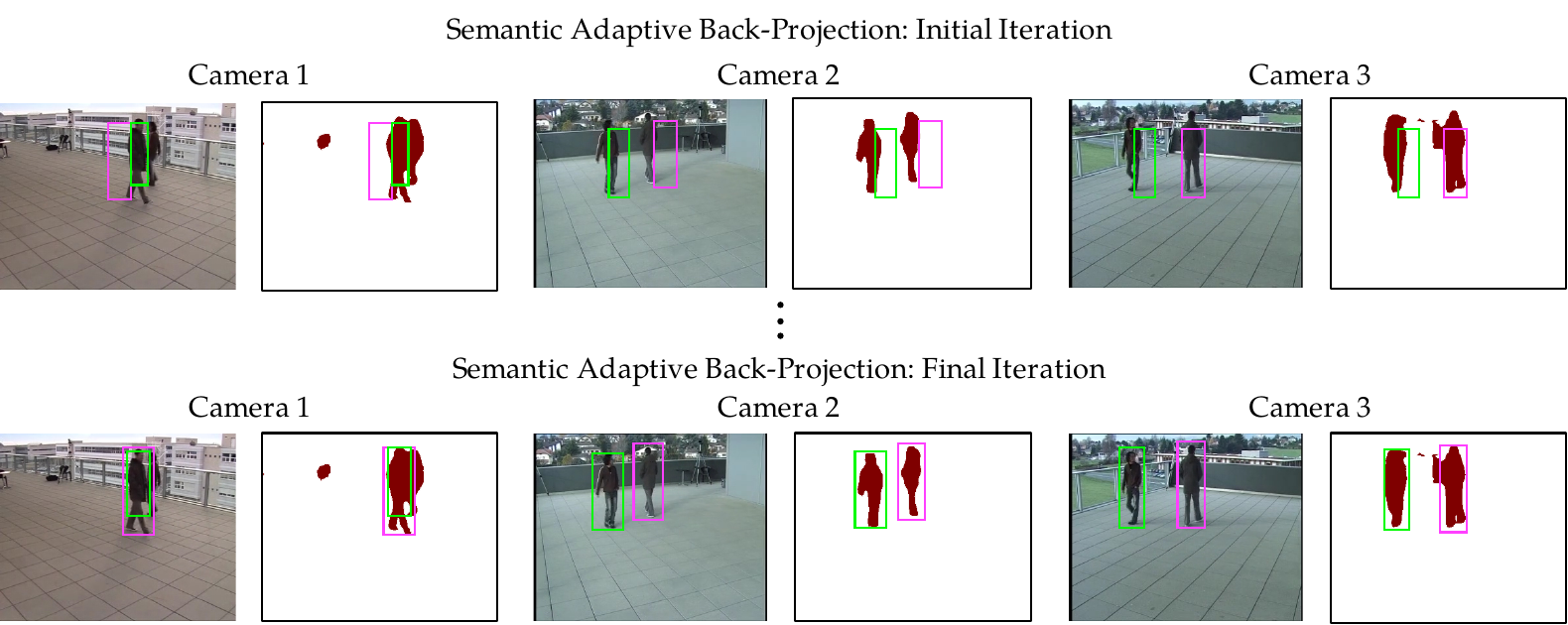}
    \caption{Ablation Studies: Semantic-Driven Back-Projection. First row: back-projected bounding boxes at the initial iteration of the optimization algorithm. Global detections obtained by the Multi-camera Detection Fusion algorithm are displaced with respect to real pedestrian when back-projected to each camera. Second row: The semantic-driven optimization algorithm correctly refines locations and heights for the bounding boxes in Camera 2 and 3. However, when semantic pedestrian cues are highly overlapped some bounding boxes might be refined to an incorrect location (Camera 1, green bounding-box). Better viewed in color.}
    \label{fig:Problems Optimization}
\end{figure*}

Table \ref{tab:Ablation Results} also shows that the complete method (Baseline + Filtering + Fusion + Back-Projection) notably improves Faster-RCNN baseline's performance, mainly in scenarios with heavy occlusions, i.e. EPFL Terrace and EPFL RLC (See Table \ref{tab:Dataset Comparison} for details). Specifically, for the EPFL Terrace dataset results are relatively increased a \(6.14\%\), a \(7.14\%\) and a \(16.90\%\) in terms of AUC, F-Score, and N-MODA respectively, whereas relative improvements are of a \(5.19\%\)---in AUC, a \(5.13\%\)---in F-Score terms---and a \(20.69\%\) in N-MODA, for the EPFL RLC dataset. 

For YOLOv3 and EfficientDet detectors a similar analysis arises. In scenarios where heavy occlusions are present\textemdash EPFL Terrace and RLC datasets, performance is increased. 
For the EPFL Terrace Dataset, relative increments of a \(2.38\%\), a \(2.29\%\) and a \(11.84\%\) are obtained when using YOLOv3 detections in terms of  AUC, F-Score and N-MODA respectively. In the case of EfficientDet, increments of a \(4.87\%\), a \(5.95\%\) and a \(16.90\%\) are obtained with respect to the same metrics. For the EPFL RLC dataset, the improvement increases to a \(6.25\%\), a \(6.41\%\) and a \(15.25\%\) for YOLOv3, whereas a \(1.21\%\), a \(3.75\%\) and a \(13.11\%\) relative increase is obtained for EfficientDet in terms of AUC, F-Score and N-MODA respectively.

For both PETS scenarios the performance of the EfficientDet mono-camera detector is saturated (97\% F-Score). The specific characteristics of this dataset: low pedestrian density over a wide space, low level of occlusions and a high point of view due to cameras being hanged up in streetlights (see Table \ref{tab:Dataset Comparison} and Figure \ref{fig:QualitativeResults}), turns it in the least complex dataset among those analysed. The generation of new false positive detections and the optimization process related problems ((Figure \ref{fig:Problems Optimization})) may lead to a slight decrease when saturated baselines are used in low complex datasets. Leaving these specific situations aside, the benefits of the proposed method are evident if one accounts for both performance indicators and qualitative results (see Table \ref{tab:Ablation Results} and Figure \ref{fig:QualitativeResults} respectively): the proposed multi-camera detection approach is able to cope with partial, severe and complete occlusions by combining detections from all the cameras through the proposed semantic-guided process leading to an increase of all the reported metrics.

Focusing specifically on the Semantic-Driven Back-Projection process, results in Figure \ref{fig:QualitativeResults} depict highly tight pedestrian bounding boxes, independently from people's height, self-occlusions and calibration problems, suggesting that the optimization process is able to automatically adapt bounding-boxes by jointly estimating pedestrian heights and world positions. Results in Table \ref{tab:Ablation Results} corroborate this observation. Semantic-Driven Back-Projection leads to a higher overlap between detections and ground-truth annotations: in terms of the N-MODP metric, the proposed method achieves relative improvements of a \(4.05\%\) for EPFL Terrace, a \(3.95\%\) for both PETS 2009 S2 L1 and PETS 2009 CC and a \(1.45\%\) for the RLC dataset when Faster-RCNN is used as the baseline detector. When YOLOv3 is used as the baseline detector, our method achieves a N-MOPD increment of a \(4.10\%\) for EPFL Terrace whereas the N-MODP metric remains stable for PETS 2009 S2 L1, PETS 2009 CC and RLC datasets, suggesting that YOLOv3 individual performance for these datasets is already heaped. A similar result arises when using EfficientDet detector which by default is highly tight to pedestrians. Results are increased only for PETS CC dataset by a \(13.63\%\) while manteined for the rest of the datasets. It is important to remind that, even tough the N-MODP metrics are sometimes slightly reduced or maintained, without the proposed Semantic-Driven Back-Projection process the back-projected bounding boxes and ground-truth would be misaligned (see Figure \ref{fig:OptimizationAlgorithm}) decreasing the performance in terms of all the accuracy metrics of the proposed method.

Nevertheless, the optimization cost function aims to maximize the 2D detections’ alignment with the semantic segmentation masks, leading to a bias towards wider pedestrians by design, a situation that may result sometimes into wrong relocations of the back-projected bounding-boxes. Figure \ref{fig:Problems Optimization} shows an example of this case: notice the erring behaviour in Camera 1 when there is an extreme overlapping.

\subsection{State-of-the-art Comparison}
\subsubsection*{Evaluation Criterion}\label{Ex2: Evaluation Criteria}
The same criterion used in the Ablation Studies applies for the Terrace, PETS and RLC datasets. However, in the Wildtrack dataset, as the ground-truth is provided via detections on the world ground plane (i.e., no bounding-boxes are provided), the evaluation criterion is different. Specifically, a detection is considered a TP if it lies at most \(r = 0.5\)m to a ground-truth annotated point \cite{web-wildtrackDataset}. This radius roughly corresponds to the average width of the human body. Due to the absence of bounding-boxes, for this dataset the Semantic-Driven Back-Projection stage is not included.

\subsubsection*{State-of-the-art Algorithms}
The following multi-camera algorithms have been selected to carry out the comparison:
\begin{itemize}
\item POM \cite{fleuret2008multicamera}. This algorithm proposes to estimate the marginal probabilities of pedestrians at every location inside an \(\mathcal{AOI}\). It is based on a preliminary background subtraction stage.
\item POM-CNN \cite{fleuret2008multicamera}. An upgraded version of POM in which the background subtraction stage is performed based on an encoder-decoder CNN architecture.
\item MvBN+HAP \cite{peng2015robust}. Relies on a multi-view Bayesian network model (MvBN) to obtain pedestrian locations on the ground plane. Detections are then refined by a Height-Adaptive Projection method (HAP) based on an optimization framework similar to the one proposed in this paper, but driven by background-subtraction cues.
\item RCNN-Projected \cite{xu2016multi}. The bottom of bounding-boxes obtained thorough per-camera CNN detectors are projected onto ground-plane, where 3D proximity is used to cluster detections.
\item Deep-Occlusion \cite{baque2017deep} is an hybrid method which combines a CNN trained on the Wildtrack dataset and a Conditional Random Fields (CRF) method to incorporate information on the geometry and calibration of the scene.
\item DeepMCD \cite{chavdarova2017deep} is an end-to-end deep learning approach based on different architectures and training scenarios:
\begin{itemize}
\item Pre-DeepMCD: a GoogleNet \cite{szegedy2015going} architecture trained on the PETS dataset.
\item Top-DeepMCD: a GoogleNet \cite{szegedy2015going} architecture trained on the Wildtrack dataset.
\item ResNet-DeepMCD: a ResNet-18 \cite{he2016deep} architecture trained on the Wildtrack dataset.
\item DenseNet-DeepMCD: a DenseNet-121 \cite{huang2017densely} architecture trained on the Wildtrack dataset.
\end{itemize}
\end{itemize}

\begin{table*}[t!]   
    \begin{centering}
    \footnotesize
    \renewcommand{\arraystretch}{2}
    \resizebox{\textwidth}{!}{
    \begin{tabular}{l c c c c c c c c c c c c}
	\cline{2-13} 
	\multicolumn{1}{l}{} & \multicolumn{12}{c}{Dataset}\tabularnewline
	\hline 
	\multirow{2}{*}{Algorithm}               &           \multicolumn{3}{c}{PETS CC}         &          \multicolumn{3}{c}{PETS S2 L1}       &      \multicolumn{3}{c}{EPFL Terrace}         &         \multicolumn{3}{c}{EPFL RLC}          \tabularnewline
	\cline{2-13}                                                                             
	                                         & F-S           & N-A           & N-P           & F-S           & N-A           & N-P           & F-S           & N-A           & N-P           & F-S           & N-A           & N-P           \tabularnewline
	\hline                                                                                   
	Faster-RCNN \cite{renNIPS15fasterrcnn}   & 0.91          & 0.85          & 0.76          & 0.91          & 0.85          & 0.76          & 0.84          & 0.71          & 0.74          & 0.78          & 0.58          & 0.69          \tabularnewline
	YOLO v3 \cite{redmon2018yolov3}          & 0.92          & 0.87          & 0.79          & 0.96          & 0.92          & 0.67          & 0.87          & 0.76          & 0.73          & 0.78          & 0.59          & 0.72          \tabularnewline
	Efficient Det \cite{tan2020efficientdet} & \textbf{0.97} & \textbf{0.94} & \textbf{0.66} & \textbf{0.97} & \textbf{0.94} & \textbf{0.67} & 0.84          & 0.71          & 0.78          & 0.80          & 0.61          & 0.72          \tabularnewline
	POM \cite{fleuret2008multicamera}        & -             & 0.70          & 0.55          & -             & 0.65          & 0.67          & -             & 0.19          & 0.56          & -             & -             & -             \tabularnewline
	MvBN + HAP \cite{peng2015robust}         & -             & 0.87          & 0.78          & -             & 0.87          & 0.76          & -             & 0.82          & 0.73          & -             & -             & -             \tabularnewline
	\hline                                                                                   
	\textbf{Ours (Faster-RCNN)}              & 0.93          & 0.88          & 0.79          & 0.93          & 0.89          & 0.79          & 0.90          & 0.83          & 0.77          & 0.81          & 0.68          & 0.70          \tabularnewline
	\textbf{Ours (YOLOv3)}                   & 0.94          & 0.88          & 0.79          & 0.92          & 0.89          & 0.67          & \textbf{0.89} & \textbf{0.85} & \textbf{0.76} & 0.83          & 0.68          & 0.72          \tabularnewline
	\textbf{Ours (Efficient Det)}            & 0.94          & 0.88          & 0.75          & 0.94          & 0.88          & 0.67          & 0.89          & 0.83          & 0.76          & \textbf{0.84} & \textbf{0.71} & \textbf{0.72} \tabularnewline
	\hline
\end{tabular}}
    \caption{State-of-the-art Comparison: Comparison with baseline (Faster-RCNN, YOLOv3 and EfficientDet) and multi-camera state-of-the-art methods non based on deep-learning (POM \cite{fleuret2008multicamera} and MvBN + HAP \cite{peng2015robust}). Bold values indicate best results in terms of N-MODA. Indicators are F-Score (F-S), N-MODA (N-A) and N-MODP (N-P).Datasets are sorted from left to right in terms of complexity according to Table \ref{tab:Dataset Comparison}.}
    \label{tab:State Of the Art Comparison}
    \end{centering}
\end{table*}

\begin{table}[t!]   
    \begin{centering}
        \renewcommand{\arraystretch}{2}
        \resizebox{\columnwidth}{!}{
        \begin{tabular}{ l c c c c c c }
            \cline{2-6} 
            {} & \multicolumn{5}{c}{EPFL Wildtrack}\tabularnewline
            \hline 
            Algorithm & Authors \(\mathcal{AOI}\) & Fine-Tuned & F-Score & N-MODA & N-MODP\tabularnewline
            \hline 
            \textbf{Deep-Occlusion} \cite{baque2017deep} & \checkmark & \checkmark & \textbf{0.86} & \textbf{0.74} & \textbf{0.53}\tabularnewline
            ResNet-DeepMCD \cite{chavdarova-et-al-2018} & \checkmark & \checkmark  & 0.83 & 0.67 & 0.64\tabularnewline
            \textbf{Ours* (EfficientDet)} & \checkmark &  & \textbf{0.81} &\textbf{0.65} & \textbf{0.63}\tabularnewline
            DenseNet-DeepMCD \cite{chavdarova-et-al-2018} & \checkmark & \checkmark  & 0.79 & 0.63 & 0.66\tabularnewline
            Top-DeepMCD \cite{chavdarova2017deep} & \checkmark & \checkmark  & 0.79 & 0.60 & 0.64\tabularnewline
            GMC-3D \cite{lima2021generalizable} & \checkmark &  & 0.78 & 0.56 & 0.67\tabularnewline
            \textbf{Ours (EfficientDet)} & &   & 0.74 & 0.48 & 0.63\tabularnewline
            \textbf{Ours (YOLOv3)} & &   & 0.71 & 0.42 & 0.60\tabularnewline
            \textbf{Ours (Faster-RCNN)} & &   & 0.69 & 0.39 & 0.55\tabularnewline
            Pre-DeepMCD \cite{chavdarova2017deep} & \checkmark &   & 0.51 & 0.33 & 0.52\tabularnewline
            POM-CNN \cite{fleuret2008multicamera} & \checkmark &   & 0.63 & 0.23 & 0.30\tabularnewline
            RCNN-Projected \cite{xu2016multi} & \checkmark &   & 0.52 & 0.11 & 0.18\tabularnewline
            \hline 
        \end{tabular}}
        \caption{State-of-the-art Comparison: Wildtrack Dataset Comparison Results. ``Authors-\(\mathcal{AOI}\)" stands for algorithm performance evaluated using the \(\mathcal{AOI}\) proposed by the authors. ``Fine-tuned" denotes that the algorithm has been explicitly trained on Wildtrack dataset. Bold values indicate best results in terms of N-MODA.}   
        \label{tab:Wildtrack Comparison}
    \end{centering}
\end{table}

\begin{figure*}[t!]
    \centering
    \includegraphics[width=1\textwidth,keepaspectratio]{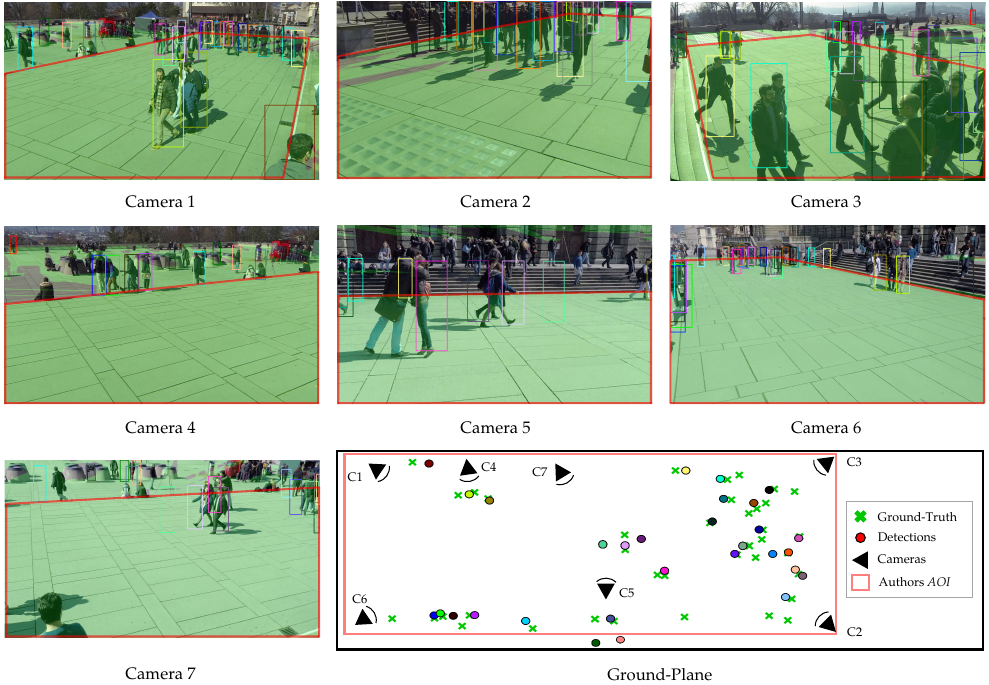}
    \caption{State-of-the-art Comparison: Qualitative results from a sample frame on Wildtrack dataset. For representation reasons, camera frames depict adapted bounding boxes via the Semantic-Driven Back-Projection stage, although this stage is not used for evaluation in the Wildtrack dataset. In addition, the Figure depicts the automatically obtained \(\mathcal{AOI}\) superimposed in green and finally, the manually annotated \(\mathcal{AOI}\) proposed by the authors \cite{web-wildtrackDataset,chavdarova-et-al-2018} (area delimited by red lines). The last image represents the cameras' positions, the obtained detections and the authors' ground-truth and \(\mathcal{AOI}\) over the ground-plane. Pedestrians are identified with different colors (one per detection) along views and ground-plane. Better viewed in color.}
    \label{fig:Wildtrack Qualitative Results}
\end{figure*}

\subsubsection*{Results}
Table \ref{tab:State Of the Art Comparison} includes performance indicators for the proposed method compared with multi-camera algorithms POM \cite{fleuret2008multicamera} and MvBN+HAP \cite{peng2015robust} on the Terrace, PETS and RLC scenarios (results for the compared methods are extracted from \cite{peng2015robust}). Table \ref{tab:Wildtrack Comparison} compares the performance of the proposed approach against deep-learning methods, some of them explicitly trained with data from the Wildtrack dataset (which we denote as \textit{Fine-Tuned}) and others trained with data from other datasets or not even trained (which we denote as not \textit{Fine-Tuned}). Performance indicators for these methods are extracted from \cite{chavdarova-et-al-2018}. In addition, qualitative results for the Wildtrack dataset are presented in Figure \ref{fig:Wildtrack Qualitative Results}, including obtained detections in camera frames, global detections on the ground-plane and the automatically computed \(\mathcal{AOI}\).

\subsubsection*{Discussion}
Results in Table \ref{tab:State Of the Art Comparison} show that the proposed approach (Baseline + Filtering + Fusion + Back-Projection), either with Faster-RCNN, YOLOv3 or EfficentDet baseline, outperforms the MvBN + HAP and the POM-CNN methods in terms of N-MODA metric. The proposed method obtains better results in terms of N-MODA which, precisely, measures detection accuracy along the whole video sequences. Best results are obtained when EfficientDet is used to extract mono-camera detections. Specifically, N-MODA is increased a \(1.21\%\) for EPFL Terrace and a \(1.14\%\) for both PETS 2009 S2 L1 and CC. Moreover, it obtains the higher performance on the heavily occluded RLC dataset. Besides, N-MODP results, i.e. the overlapping between detections and ground-truth, are better than those obtained by the HAP method \cite{peng2015robust}. This suggests that our use of semantic segmentation masks instead of foreground masks (HAP method) benefits the optimization process. Relative increments in N-MODP performance of a \(5.48\%\) for EPFL Terrace, a \(3.95\%\) for PETS 2009 S2 L1 and a \(1.28\%\) for PETS 2009 CC support this assumption.

Presented results in Table \ref{tab:State Of the Art Comparison} suggest that the proposed method is able to obtain reliable pedestrian detections in a variety of scenarios with a variety of pedestrian detection algorithms in terms of performance.

Finally, results on the Wildtrack dataset (Table \ref{tab:Wildtrack Comparison}), indicate that the proposed method, operating on detections from a Faster-RCNN, a YOLOv3 or a EfficientDet model, is able to outperform deep-learning approaches that have not been specifically trained using Wildtrack data and use manually annotated detection constrains. Our method, using EfficientDet detections, improves a \(45.45\%\) respect to Pre-DeepMCD---the second ranked, which is an end-to-end deep learning architecture trained on the PETS dataset. However, algorithms explicitly trained on data from the Wildtrack dataset, i.e., DenseNet-DeepMCD, ResNet-DeepMCD, Top-DeepMCD, and Deep-Occlusion, outperform the proposed method, in our opinion for two main reasons:

\begin{itemize}
  \item First, the qualitative results presented in Figure \ref{fig:Wildtrack Qualitative Results} suggest that results in Table \ref{tab:Wildtrack Comparison} are highly biased by the authors' manually annotated area. The proposed method obtains a broader \(\mathcal{AOI}\) (Figure \ref{fig:Wildtrack Qualitative Results}, green area) than the one provided by the authors (Figure \ref{fig:Wildtrack Qualitative Results}, red area). Although the automatically obtained \(\mathcal{AOI}\) seems to be better fitted to the ground floor in the scene than the manually annotated one, the performance of our method decreases because ground-truth data is reported only on the manually annotated area. Thereby, our true positive detections out of this area result in false positives in the statistics (see Figure \ref{fig:Wildtrack Qualitative Results}, cameras 1 and 4).
  
  \item Second, they learn their occlusion modeling and their inference ground occupancy probabilistic models specifically on the Wildtrack scenario using samples from the dataset. This training, as any fine-tuning procedure in Deep Neural Networks, is highly effective, as indicated by the increase in performance resulting from the use of the same architecture but adapted for the Wildtrack scenario (compare results of Pre-DeepMCD and Top-DeepMCD). This training requires the use of human-annotated detections in each scenario, hindering the scalability of these solutions and its application to the real world. The proposed approach, on the other hand, performs equally without the need of being adapted for every target scenario reported in this paper. 
\end{itemize}

Respect to the first issue, i.e. the effect of using the proposed automatically extracted \(\mathcal{AOI}\) instead of the one provided by the dataset and used by the rest of the methods, in order to obtain a fairer comparison, we have included in Table \ref{tab:Wildtrack Comparison} the result of the proposed method evaluated on the authors' \(\mathcal{AOI}\) using our top ranked method, i.e. using EfficientDet baseline. As it can be observed, when using the authors' \(\mathcal{AOI}\) for evaluation, the proposed method outperforms TOP-DeepMCD by a \(8.33\%\) and DenseNet-DeepMCD by a \(3.17\%\) ranking the third best method on the Wildtrack dataset without requiring a dataset specific fine-tunning stage as the two above it. In addition, performance with respect to GMC-3D \cite{lima2021generalizable}, which replicates the previous version of the proposed method with the addition of Person Re-Identification features is increased a \(17.85 \%\).

On average, and contrary to state-of-the-art approaches, the proposed method adapts to different target scenarios without needing a separate training stage for each situation, with the consequent reduction of computational resources and time, and neither requiring a manually annotated area of interest.


\section{Conclusions} \label{Conclusions}
This paper describes a novel approach to perform pedestrian detection in a multi-camera recorded scenario. First, the adapted strategies for the temporal and spatial aggregation of semantic cues, along with homography projections, are used to obtain an estimation of the ground-plane. Through this process, a broader, accurate and role-annotated Area of Interest (\(\mathcal{AOI}\)) is automatically defined. Per-camera detections, obtained by a state-of-the-art detector, are projected to the reference plane, and those laying outside the obtained \(\mathcal{AOI}\) are filtered-out. A fusion approach based on creating connected components on a graph representation of the detections is used to combine per-camera detections yielding global pedestrian detection. Then, a Semantic-Driven Back-Projection method handles occlusions and uses semantic cues to globally refine the location and size of the back-projected detections by aggregating information from all the cameras. Results on a broad set of scenarios confirm that the method outperforms every other compared multi-camera not deep-learning method and also every deep-learning method not adapted to the target dataset, even with different baseline algorithms. The proposed method performs close to scenario-tailored methods, but without their training stage, which highly hinders their straight use in new scenarios. In overall, results suggest that the proposed approach is able to obtain accurate, robust, tight-to-object and generic pedestrian detection in varied scenarios, included crowded ones.

\begin{acknowledgements}
This study has been partially supported by the Spanish Government through its TEC2017-88169-R MobiNetVideo project. 
\end{acknowledgements}

%
%

\bibliographystyle{spmpsci}      
\bibliography{mybibfile.bib}   




\end{document}